\documentstyle[epic,bezier,acl]{article}

\title{\vspace{-14mm}\LARGE\bf TEXT SEGMENTATION \\
                               BASED ON SIMILARITY BETWEEN WORDS}

\author{  Hideki \ Kozima \\
{\large Course in Computer Science and Information Mathematics,   } \\
{\large Graduate School, \ \ University of Electro-Communications } \\
{\large 1--5--1, Chofugaoka, \ Chofu,                             } \\
{\large Tokyo 182, \ Japan                                        } \\
{\large ({\tt xkozima@phaeton.cs.uec.ac.jp})                      } }

\newlength{\lma}
\newlength{\lmb}
\newlength{\lmc}

\newcounter{pc}

\newenvironment{itemizing}{
  \vspace{2mm}
  \begin{list}
    {$\bullet$}{\setlength{\leftmargin}{\lmb}
                \setlength{\topsep}{0mm}
                \setlength{\itemsep}{1mm}
                \setlength{\parsep}{0mm} }}{
  \end{list}
  \vspace{2mm} }

\newenvironment{references}{
  \vspace{2mm}
  \begin{list}
    {}{\setlength{\leftmargin}{5mm}
       \setlength{\topsep}{0mm}
       \setlength{\itemsep}{1mm}
       \setlength{\parsep}{0mm} 
       \setlength{\itemindent}{-5mm} }}{
  \end{list}
  \vspace{2mm} }

\begin{document}

\maketitle


\abstract{This paper proposes a new indicator of 
text structure, called the lexical cohesion profile
(LCP), which locates segment boundaries in a text.
A text segment is a coherent scene; the words in a
segment are linked together via lexical cohesion
relations.  LCP records mutual similarity of words
in a sequence of text.  The similarity of words,
which represents their cohesiveness, is computed
using a semantic network.  Comparison with the text
segments marked by a number of subjects shows that
LCP closely correlates with the human judgments.
LCP may provide valuable information for resolving
anaphora and ellipsis.}


\section{INTRODUCTION}

A text is not just a sequence of words, but it has
coherent structure.  The meaning of each word can
not be determined until it is placed in the
structure of the text.  Recognizing the structure of
text is an essential task in text understanding,
especially in resolving anaphora and ellipsis.

One of the constituents of the text structure is a
text segment.  A text segment, whether or not it is
explicitly marked, as are sentences and paragraphs,
is defined as a sequence of clauses or sentences
that display local coherence.  It resembles a scene
in a movie, which describes the same objects in the
same situation.

This paper proposes an indicator, called the lexical
cohesion profile (LCP), which locates segment
boundaries in a narrative text.  LCP is a record of
lexical cohesiveness of words in a sequence of text.
Lexical cohesiveness is defined as word similarity
(Kozima and Furugori, 1993) computed by spreading
activation on a semantic network.  Hills and valleys
of LCP closely correlate with changing of segments.


\section{SEGMENTS AND COHERENCE}

Several methods to capture segment boundaries have
been proposed in the studies of text structure.  For
example, cue phrases play an important role in
signaling segment changes. (Grosz and Sidner, 1986)
\ However, such clues are not directly based on
coherence which forms the clauses or sentences into
a segment.

Youmans (1991) proposed VMP (vocabulary management
profile) as an indicator of segment boundaries.  VMP
is a record of the number of new vocabulary terms
introduced in an interval of text.  However, VMP
does not work well on a high-density text.  The
reason is that coherence of a segment should be
determined not only by reiteration of words but also
by lexical cohesion.

Morris and Hirst (1991) used Roget's thesaurus to 
determine whether or not two words have lexical 
cohesion.  Their method can capture almost all the 
types of lexical cohesion, e.g. systematic and 
non-systematic semantic relation.  However it does 
not deal with strength of cohesiveness which 
suggests the degree of contribution to coherence 
of the segment.

\subsection{Computing Lexical Cohesion}

Kozima and Furugori (1993) defined lexical 
cohesiveness as semantic similarity between words, 
and proposed a method for measuring it.  
Similarity between words is computed by spreading 
activation on a semantic network which is 
systematically constructed from an English 
dictionary (LDOCE).

The similarity $\sigma(w,w') \!\in\! [0,1]$ 
between words $w,w'$ is computed in the following 
way: (1) produce an activated pattern by activating 
the node $w$; (2) observe activity of the node $w'$ 
in the activated pattern.  The following examples 
suggest the feature of the similarity $\sigma$:
\begin{tabbing}
\hspace{8mm}\=\hspace{22mm}\=\hspace{22mm}\=\hspace{5mm}\=\kill
\> $\sigma$ ({\tt cat},      \> {\tt pet})        \> = \> 0.133722 , \\
\> $\sigma$ ({\tt cat},      \> {\tt hat})        \> = \> 0.001784 , \\
\> $\sigma$ ({\tt waiter},   \> {\tt restaurant}) \> = \> 0.175699 , \\
\> $\sigma$ ({\tt painter},  \> {\tt restaurant}) \> = \> 0.006260 .
\end{tabbing}

The similarity $\sigma$ depends on the significance
$s(w) \!\in\! [0,1]$, i.e. normalized information of
the word $w$ in West's corpus (1953).  For example:
\begin{tabbing}
\hspace{8mm}\=\hspace{34mm}\=\kill
\> s({\tt red}) = 0.500955 , \> s({\tt and}) = 0.254294 .
\end{tabbing}
The following examples show the relationship 
between the word significance and the similarity:
\begin{tabbing}
\hspace{8mm}\=\hspace{22mm}\=\hspace{22mm}\=\hspace{5mm}\=\kill
\> $\sigma$ ({\tt waiter},   \> {\tt waiter})     \> = \> 0.596803 , \\
\> $\sigma$ ({\tt red},      \> {\tt blood})      \> = \> 0.111443 , \\
\> $\sigma$ ({\tt of},       \> {\tt blood})      \> = \> 0.001041 .
\end{tabbing}


\section{LEXICAL COHESION PROFILE}

LCP of the text $T \!=\! \{w_1, \!\cdots\!, w_N\}$ 
is a sequence $\{c(S_1), \!\cdots\!, c(S_N)\}$ of 
lexical cohesiveness $c(S_i)$.  $S_i$ is the word 
list which can be seen through a fixed-width 
window centered on the $i$-th word of $T$:
\begin{tabbing}\hspace{5mm}\=\hspace{3mm}\=\hspace{3mm}\=
               \hspace{20mm}\=\kill
\> $S_i$=$\{w_{l},w_{l+1},\cdots,w_{i-1},w_{i},
            w_{i+1},\cdots,w_{r-1},w_{r}\}$, \\
\> \> $l$ \> = $i \!-\! \Delta$
      \> (if $i \!\leq\! \Delta$, then $l \!=\! 1$),\\
\> \> $r$ \> = $i \!+\! \Delta$
      \> (if $i \!>\! N \!-\! \Delta$, then $r \!=\! N$).
\end{tabbing}
LCP treats the text $T$ as a word list without any 
punctuation or paragraph boundaries.

\begin{figure}
\begin{center}
\\
\vspace{-3mm}
{\bf Figure 1.} \  An activated pattern of a word list \\
(produced from \{{\tt red}, {\tt alcoholic}, {\tt drink}\}).
\vspace{-3mm}
\end{center}\end{figure}

\subsection{Cohesiveness of a Word List}

Lexical cohesiveness $c(S_i)$ of the word list 
$S_i$ is defined as follows:
\vspace{-1mm}
\begin{center}
   $c(S_i) = \sum_{w \in S_i}
             s(w) \!\cdot\! a(P(S_i),w)$ ,
\end{center}
\vspace{-1mm}
where $a(P(S_i),w)$ is the activity value of the 
node $w$ in the activated pattern $P(S_i)$.  
$P(S_i)$ is produced by activating each node $w 
\!\in\! S_i$ with strength $s(w)^2 / \sum s(w)$.
Figure 1 shows a sample pattern of \{{\tt red}, 
{\tt alcoholic}, {\tt drink}\}.  (Note that it has 
highly activated nodes like {\tt bottle} and {\tt 
wine}.)

The definition of $c(S_i)$ above expresses that 
$c(S_i)$ represents semantic homogeneity of $S_i$, 
since $P(S_i)$ represents the average meaning of 
$w \!\in\! S_i$.  For example:
\begin{tabbing}
\hspace{5mm}\=\hspace{2mm}\=\hspace{2mm}\=\hspace{55mm}\=\hspace{5mm}\=\kill
\>$c$\>(\> {\tt "Molly saw a cat.  It was her family} \\
\>   \> \> {\tt pet.  She wished to keep a lion."} \\
\>   \> = 0.403239 \ \ (cohesive), 
\end{tabbing}\begin{tabbing}
\hspace{5mm}\=\hspace{2mm}\=\hspace{2mm}\=\hspace{55mm}\=\hspace{5mm}\=\kill
\>$c$\>(\> {\tt "There is no one but me.  Put on }\\
\>   \> \> {\tt your clothes.  I can not walk more."} \\
\>   \> = 0.235462 \ \ (not cohesive). 
\end{tabbing}

\begin{figure}[t]\begin{center}
\\
{\bf Figure 2.} \ 
\vspace{-2.5mm}
\begin{minipage}[t]{40mm}
Correlation between LCP\\
and text segments.
\vspace{-1mm}
\end{minipage}\end{center}\end{figure}

\begin{figure}
\begin{center}
%
\\
\vspace{-2mm}
\ \ \ {\bf Figure 3.} \ An example of LCP \\
\ \ (using rectangular window of $\Delta$=25)
\vspace{-2mm}
\end{center}\end{figure}

\subsection{LCP and Its Feature}

A graph of LCP, which plots $c(S_i)$ at the text 
position $i$, indicates changing of segments:
\begin{itemizing}
\item If $S_i$ is inside a segment, it tends to be 
cohesive and makes $c(S_i)$ high.
\item If $S_i$ is crossing a segment boundary, it 
tends to semantically vary and makes $c(S_i)$ low.
\end{itemizing}
As shown in Figure 2, the segment boundaries can 
be detected by the valleys (minimum points) of LCP.

The LCP, shown in Figure 3, has large hills and
valleys, and also meaningless noise.  The graph is
so complicated that one can not easily determine
which valley should be considered as a segment
boundary.

\begin{figure*}[t]
\begin{center}
\\
\vspace{-3mm}
\ \ \ {\bf Figure 4.} \ Correlation between LCP and segment boundaries.
\vspace{-3mm}
\end{center}\end{figure*}

The shape of the window, which defines weight of 
words in it for pattern production, makes LCP 
smooth.  Experiments on several window shapes 
(e.g. triangle window, etc.) shows that Hanning 
window is best for clarifying the macroscopic 
features of LCP.

The width of the window also has effect on the 
macroscopic features of LCP, especially on 
separability of segments.  Experiments on several 
window widths ($\Delta \!=\! 5 \sim 60$) reveals 
that the Hanning window of $\Delta \!=\! 25$ gives 
the best correlation between LCP and segments.


\section{VERIFICATION OF LCP}

This section inspects the correlation between LCP 
and segment boundaries perceived by the human 
judgments. The curve of Figure 4 shows the LCP of  
the simplified version of O.Henry's ``Springtime 
\`a la Carte'' (Thornley, 1960).  The solid bars 
represent the histogram of segment boundaries 
reported by 16 subjects who read the text without 
paragraph structure.

It is clear that the valleys of the LCP correspond 
mostly to the dominant segment boundaries.  For 
example, the clear valley at $i \!=\! 110$ exactly 
corresponds to the dominant segment boundary (and 
also to the paragraph boundary shown as a dotted 
line).

Note that LCP can detect segment changing of a 
text regardless of its paragraph structure.  For 
example, $i \!=\! 156$ is a paragraph boundary, 
but neither a valley of the LCP nor a segment 
boundary; \ $i \!=\! 236$ is both a segment 
boundary and approximately a valley of the LCP, 
but not a paragraph boundary.

However, some valleys of the LCP do not exactly 
correspond to segment boundaries.  For example, 
the valley near $i \!=\! 450$ disagree with the 
segment boundary at $i \!=\! 465$.  The reason is 
that lexical cohesion can not cover all aspect of 
coherence of a segment; \ an incoherent piece of 
text can be lexically cohesive.


\section{CONCLUSION}

This paper proposed LCP, an indicator of segment 
changing, which concentrates on lexical cohesion 
of a text segment.  The experiment proved that LCP 
closely correlate with the segment boundaries 
captured by the human judgments, and that lexical 
cohesion plays main role in forming a sequence of 
words into segments.

Text segmentation described here provides basic 
information for text understanding:
\begin{itemizing}
\item Resolving anaphora and ellipsis: \\
Segment boundaries provide valuable restriction 
for determination of the referents.
\item Analyzing text structure: \\
Segment boundaries can be considered as segment 
switching (push and pop) in hierarchical structure 
of text.
\end{itemizing}
The segmentation can be applied also to text 
summarizing. (Consider a list of average meaning 
of segments.)

In future research, the author needs to examine
validity of LCP for other genres --- Hearst (1993)
segments expository texts.  Incorporating other
clues (e.g. cue phrases, tense and aspect, etc.) is
also needed to make this segmentation method more
robust.


\section{ACKNOWLEDGMENTS}

The author is very grateful to Dr.~Teiji Furugori,
University of Electro-Communications, for his
insightful suggestions and comments on this work.



\end{document}